\pdfoutput=1
\documentclass[11pt]{article}

\usepackage[preprint]{acl}

\usepackage{times}
\usepackage{latexsym}

\usepackage[T1]{fontenc}
\usepackage[utf8]{inputenc}

\usepackage{microtype}

\usepackage{inconsolata}

\usepackage{amssymb}
\usepackage{amsmath}
\usepackage{amsfonts}
\usepackage{graphicx}
\usepackage{threeparttable}
\usepackage{booktabs}
\usepackage[whole]{bxcjkjatype} 
\usepackage{subfig}
\usepackage{xspace}

\newcommand{\clippings}{\texttt{CLIPPINGS}\xspace}

\begin{document}

%%%%%%%%% TITLE
\title{Record Linkage with Multimodal Contrastive Learning} 

\author
{Abhishek Arora$^{1}$, Xinmei Yang$^{1}$,  Shao-Yu Jheng$^{1}$, Melissa Dell$^{1, 2\ast}$ \\
\normalsize{$^{1}$Harvard University; Cambridge, MA, USA.}\\
\normalsize{$^{2}$National Bureau of Economic Research; Cambridge, MA, USA.}\\
\normalsize{$^{\ast}$Corresponding author:  melissadell@fas.harvard.edu.}
}

\maketitle

%%%%%%%%% ABSTRACT
\begin{abstract}
  Many applications require linking individuals, firms, or locations across datasets. Most widely used methods, especially in social science, do not employ deep learning, with record linkage commonly approached using string matching techniques. Moreover, existing methods do not exploit the inherently multimodal nature of documents. In historical record linkage applications, documents are typically noisily transcribed by optical character recognition (OCR). Linkage with just OCR'ed texts may fail due to noise, whereas linkage with just image crops may also fail because vision models lack language understanding (\textit{e.g.,} of abbreviations or other different ways of writing firm names). To leverage multimodal learning, this study develops \clippings (\textbf{C}ontrastively \textbf{LI}nking \textbf{P}ooled \textbf{P}re-trained Embedd\textbf{ings}). \clippings aligns symmetric vision and language bi-encoders, through contrastive language-image pre-training on document images and their corresponding OCR'ed texts. It then contrastively learns a metric space where the pooled image-text embedding for a given instance is close to embeddings in the same class (\textit{e.g.}, the same firm or location) and distant from embeddings of a different class. Data are linked by treating linkage as a nearest neighbor retrieval problem with the multimodal embeddings. \clippings outperforms widely used string matching methods by a wide margin in linking mid-20th century Japanese firms across financial documents. A purely self-supervised model - trained only by aligning the embeddings for the image crop of a firm name and its corresponding OCR'ed text - also outperforms popular string matching methods. Fascinatingly, a multimodally pre-trained vision-only encoder outperforms a unimodally pre-trained vision-only encoder, illustrating the power of multimodal pre-training even if only one modality is available for linking at inference time. 
 
\end{abstract}

%%%%%%%%% BODY TEXT
\section{Introduction}

Linking information across sources is fundamental to many analyses. For example, researchers and businesses frequently link individuals or firms across censuses and company records, governments de-duplicate benefit or voter rolls across locations, and analysts seek to identify how information from the same source spreads through media. In large swathes of the relevant literatures, deep neural methods have made few inroads. For example, a recent comprehensive review of the computer science, social science, and statistical record linkage literatures in \textit{Science Advances} \cite{binette2022almost} concludes that other methods are preferred over deep neural models for record linkage in structured data. This contrasts starkly with the literature on disambiguating entities in unstructured texts, where transformer language models overwhelmingly predominate, \textit{e.g.} \citet{wu2019scalable, de2020autoregressive, yamada2020luke}. This distinction is explained by the premise that entity linkage with unstructured texts can leverage the power of transfer learning from pre-trained large language models, whereas entity linkage in structured text databases cannot \cite{binette2022almost}. 

\begin{figure}[t]
\centering
\includegraphics[width=0.45\textwidth]{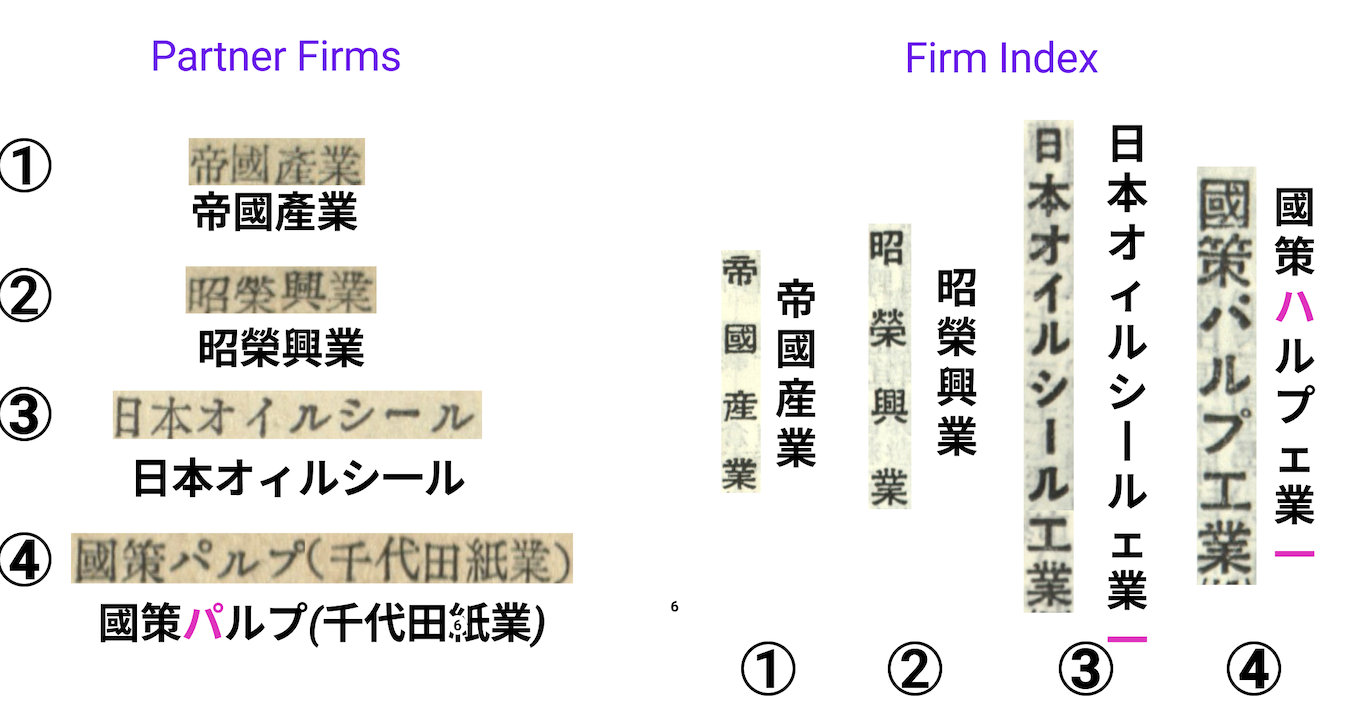}
\caption{This figure shows representative Japanese firm records, taken from a publication on trading partners (left) and a firm index (right). The text orientation of these publications differs. The numbering denotes linked firms. Pink text denotes OCR errors.} \label{fig:data_japan}
\end{figure}

This study focuses on record linkage in historical documents. We show that leveraging the power of deep learning can significantly improve the accuracy of linking firms across data sources, a common record linkage application, using historical Japanese data as a test case.
Figure \ref{fig:data_japan} shows Japanese firm level records on supply chains \cite{pr} (left) that need to be linked with a large firm level directory \cite{teikoku} (right) with rich information; the texts are written with different orientations. 
Each firm in the dataset is represented by its localized image crop and by its OCR. The image crops are useful for record linkage because the OCR contains noise; even small OCR errors can destroy significant information when matching entities with similar names. The text is useful because there are visually distinctive ways of writing the same firm's name that are discernible through language understanding. 
We focus on supply chains as they are fundamental to the transmission of economic shocks \cite{acemoglu2016networks,acemoglu2012network}, agglomeration \cite{ellison2010causes}, and economic development \cite{hirschman1958strategy, myrdal1957economic, rasmussen1956studies, bartelme2015linkages, lane2022manufacturing}, but their role in long-run economic development has been difficult to study due to the challenges of accurately linking large-scale historical records. 

To leverage both the visual and textual information illustrated in Figure \ref{fig:data_japan}, the study develops \clippings (\textbf{C}ontrastively \textbf{LI}nking \textbf{P}ooled \textbf{P}re-trained Embedd\textbf{ings}). \clippings employs end-to-end training of symmetric vision and language bi-encoders to learn a metric space where the pooled image-text representation for an instance is close to representations in the same class and distant from those in different classes. 

To train \clippings, we implement self-supervised contrastive language-image pre-training on image crop-OCR pairs, starting with a pre-trained Japanese CLIP encoder \cite{japanese-clip}. Language-image pre-training ensures that image crops are aligned with their corresponding texts, even in the presence of OCR noise.  
%The self-supervised model can be used for record linkage off-the-shelf, or 

\clippings can then be further tuned by using a contrastive loss and paired data - \textit{e.g.}, pairing records on the left and right of Figure \ref{fig:data_japan} - to align pooled image-text embeddings. These labeled data were created by the study authors. To tune the model, we use a loss function that aligns image-image, text-text, and image-text embeddings of paired data. Supervised contrastive loss \cite{khosla2020supervised} is a special case of this loss when the problem is reduced to a single modality.
Paired training data include a modest number of human gold standard annotations, in which a human determined whether records referred to the same firm. We also generated training data synthetically. 
Specifically, we rendered a list of common Japanese words as images using different fonts, applied image augmentations, and fed the resulting images to OCR. For the same word, this produced varying views of the image, as well as different views of the text due to varying OCR errors induced by the image augmentations. 

At inference time, entities are linked by retrieving the nearest neighbor(s) of queries in the target dataset (in our example, queries come from the supplier lists and the target dataset is the firm directory with richer information about the firms). The contrastive framework can flexibly incorporate blocking on database characteristics - common in record linkage - by using a type-specific loss \cite{leszczynski2022tabi}. 

%Standard powerful methods for image-text classification - used for applications such as hateful memes \cite{kiela2020hateful} or product descriptions \cite{liu2021cma, kiela2019supervised} - use a supervised classification objective that cannot be applied to problems like record linkage that have a very large number (potentially many millions) of classes, have classes that are unknown ex ante, or have classes that are constantly evolving as databases are updated. Much as in unimodal applications, such settings can be tackled by learning a metric space where classes can be assigned using kNN classification or clustering. 

\clippings significantly outperforms string matching methods - which predominate in the record linkage literature - as well as unimodal deep neural methods. 
When \clippings is trained on a modest number of linked firms using supervised contrastive learning, it achieves a 94.5\% record linkage accuracy, compared to a string matching maximum accuracy of 73.1\% achieved with custom OCR. %
When we forgo supervised contrastive training - using only the self-supervised language-image pre-training on the image-OCR pairs - this also outperforms string matching, with an accuracy of 84.9\%. 
When we use only the \clippings vision encoder, accuracy is higher than when using a unimodally trained vision encoder, aligning records that require some language understanding. The multimodal pre-training allows the vision encoder to learn some language understanding.
The \clippings models and training data are publicly released (CC-BY license).
%\clippings does not model cross-modal attention, because in record linkage applications where the inputs are an image of a text and the corresponding OCR, cross-modal attention is unlikely to lead to significantly richer representations. If desired, though, it would be straightforward to extend the framework to include cross-modal attention.

The rest of this study is organized as follows: 
Section \ref{lit} discusses the literature, and
Section \ref{methods} describes the \clippings architecture, and Section \ref{linkage} applies \clippings to the firm record linkage problem. Finally, Section \ref{limits} outlines limitations, and Section \ref{ethics} discusses ethical considerations. 

\section{Literature} \label{lit}

\textbf{Contrastive learning}:
\clippings learns a metric space for pooled image-text representations that can be applied to linkage problems even when the number of classes is extremely large, unknown ex ante, or constantly changes as databases are updated. It is reminiscent of a variety of unimodal bi-encoder applications, such as semantic similarity \cite{reimers2019sentence}, passage retrieval \cite{karpukhin2020dense}, entity disambiguation \cite{wu2019scalable} and co-reference resolution \cite{hsu2022contrastive} in unstructured text. 

To create pooled image-text representations, it is necessary to have an aligned space. 
Contrastive Language-Image Pre-training (CLIP) \cite{radford2021learning} contrastively trained aligned text and image encoders using 400 million image-caption pairs.
\clippings begins with pre-trained Japanese \cite{japanese-clip} CLIP image and text encoders. Japanese CLIP was trained with the standard CLIP \cite{radford2021learning} loss but used a BERT-based text encoder and the vision transformer was initialized by weights from the AugReg ViT-B/16 model \cite{augreg}.

\textbf{Record Linkage:}
A variety of disciplines - including computer science, statistics, database management, economics, and political science - have made extensive methodological contributions to record linkage, alternatively referred to as entity resolution, fuzzy matching, approximate dictionary matching, and string matching. 

Within this sprawling landscape, the literatures on entity resolution in structured databases \cite{binette2022almost} versus  natural language text (\textit{e.g.} \cite{wu2019scalable, de2020autoregressive, yamada2020luke}) remain largely divorced.
In the literature on record linkage in structured data - which emphasizes linking noisy text fields that contain information such as individual names, firm names, organizations, or locations -
edit distance metrics are commonly used, \textit{e.g.} \cite{levenshtein1966binary, jaro1989advances, winkler1990string}.
Another widespread approach computes the cosine similarity between $n$-gram representations of strings, where $n$-grams are defined as all substrings of size $n$ in a string \cite{okazaki2010simple}.

\begin{figure*}
    \centering
    \includegraphics[width=\linewidth]{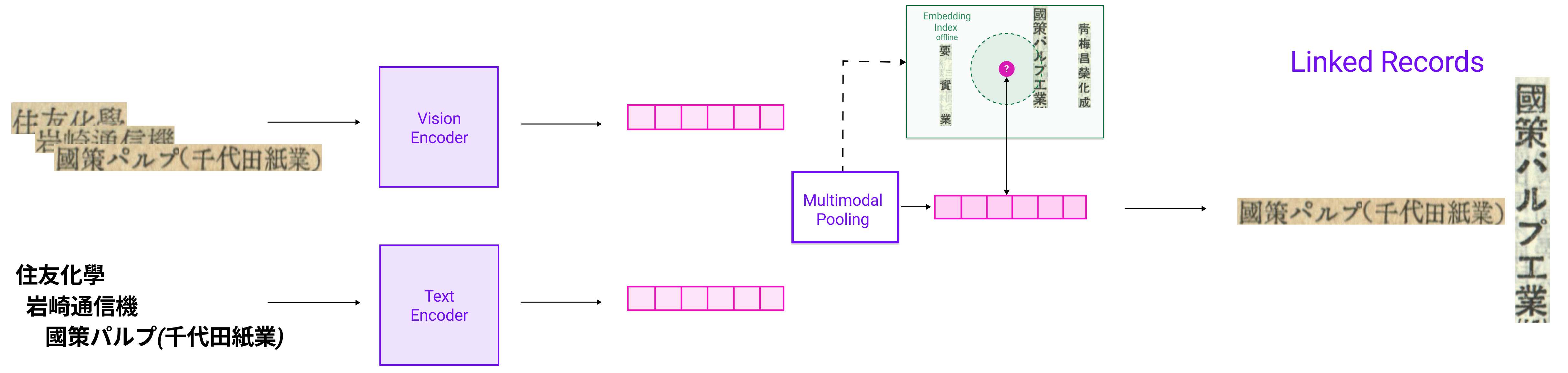}
    \caption{Model Architecture: This figure illustrates the \texttt{CLIPPINGS} model architecture.}
    \label{fig:arch}
    \vspace{-4mm}
  \end{figure*}

A recent literature on industry applications, focused on matching across e-commerce datasets, shows the promise of transformer large language models (LLMs) for improving record linkage in structured datasets \cite{li2020deep, joshi2021relink, brunner2020entity, zhou2022relation, peeters2023using, tang2022generic}. Yet these methods have not yet made widespread inroads in social science applications, with rule-based methods continuing to overwhelmingly predominate (\textit{e.g.}, see reviews by \citet{binette2022almost, abramitzky2021automated}).
Because labeled record linkage datasets are very small compared to the massive corpora used for training transformer models from scratch, a comprehensive 2022 review of the record linkage literature in \textit{Science Advances} \cite{binette2022almost} concludes that deep neural models are unlikely to be applicable to entity resolution using structured data. Constructing training data for record linkage is indeed highly labor intensive, but much of the knowledge needed to improve record linkage is plausibly already encapsulated in pre-trained image and language encoders such as 
CLIP \cite{radford2021learning}, or can be gleaned from the further self-supervised language-image pre-training pursued in this study. 

In their simplest form, approximate string matching methods simply count the required number of edits (insertions, deletions, substitutions) to transform one string into another. In practice, not all substitutions are equally probable, leading to efforts to construct rule-based lists that adjust the costs of substitutions. 
For example, the fuzzychinese \cite{fuzzychinese} package uses strokes or radicals as the fundamental unit for $n$-grams substring representations of entities, where these strokes and radicals are drawn from an external database \cite{chaizi} covering a subset of the CJK script. Alternatively, the masala merge package \cite{masala} adjusts Levenshtein distance \cite{levenshtein1966binary} to impose smaller penalties for common alternative spellings in Hindi.  Soundex, first developed in 1918 - together with the updated 1970 New York State Identification and Intelligence System (NYSIIS) \cite{silbert1970world} - account for the fact that similar sounding substitutions are more likely since census enumerators misspelled names according to their sound. These remain a bedrock for record linkage in historical U.S. census data \cite{abramitzky2021automated}.

Such rule-based methods may perform well in the contexts to which they are tailored. However, they can be brittle and are labor-intensive to extend to new settings, due to the use of hand-crafted features. This heavily skews linked datasets in social science towards a few high resource settings for which existing methods have been tailored, in turn skewing downstream knowledge. Even in high resource settings, low accuracy in some applications can require extensive human intervention in the matching process to achieve the desired accuracy \cite{bailey2020well}, limiting the scale of problems. 

%An additional limitation of string matching is that it is unimodal, despite a growing emphasis on processing documents with multimodal models or OCR-free vision-only models \cite{park2019cord, guo2019eaten, mathew2021docvqa, rust2022language, xu2020layoutlmv2, kim2022ocr, java2023one, appalaraju2021docformer, li2021selfdoc, huang2023language}.

There have also been some efforts to estimate machine learning models for record linkage. For example, \cite{ventura2015seeing} use a random forest classifier trained on labeled data to disambiguate authors of U.S. patents, applying clustering to the resulting dissimilarity scores to enforce transitivity. Very recently, the LinkTransformer package for using large language models to link structured data has been released \cite{arora2023linktransformer}. 
This package cites the challenges of using highly technical existing codebases as an impediment to the use of deep learning for record linkage in social science, and is designed to be user friendly for social scientists who lack familiarity with deep learning frameworks.
The \clippings codebase is similarly designed to be intuitive to social science users. 

\section{Model Architecture} \label{methods}

Figure \ref{fig:arch} shows the \clippings architecture. 
We begin with pre-trained, aligned image and text embeddings from Japanese language CLIP \cite{japanese-clip}, which was trained on image-caption pairs using captions machine-translated into Japanese. 
We continue self-supervised pre-training of the Japanese CLIP vision and text encoders on image crop-OCR pairs.

%Text crops often have highly diverse aspect ratios, appearing as thin vertically or horizontally oriented rectangles, while vision encoders are typically trained on square images. Center cropping would remove crucial information, and resizing can distort the image due to the mismatch in aspect ratios. To maintain the aspect ratio, we pad the rectangular crop with the median value of the border pixel, ensuring that the text region remains centered.

We then learn a metric space where the pooled image-text representation for a given instance is close to representations (embeddings) in the same class (\textit{e.g.}, the same firm) and distant from representations in different classes. 
The model is initialized with the encoders obtained through the self-supervised pre-training described above.
We use a supervised contrastive loss \cite{khosla2020supervised} on the pooled representations: 

\begin{equation*}
    -\sum_{i \in \mathcal{B}} \frac{1}{|\mathcal{P}(i)|} \sum_{k \in \mathcal{P}(i)} \log \frac{\exp \left(\tau \boldsymbol({z}_i)^T ({z}_k)\right)}{\sum_{j \in \mathcal{B}} \exp \left(\tau \boldsymbol({z}_i)^T \boldsymbol({z}_j)\right)}
\end{equation*}

where  $z_i = \dfrac{f(x_i) + g(t_i)}{2}$ is the mean of the image and text embeddings for instance $i$. 
$\mathcal{B}$ denotes the batch and $\tau$ is a temperature hyperparameter. This loss incentivizes alignment of image-image, text-text, image-text, and text-image representations across positive instances. It has the flavor of combining contrastive learning on text, contrastive learning on images, and UniCL \cite{yang2022unified}, which has a bi-directional image-text and text-image similarity objective. 
\clippings does not model cross-modal attention, because in record linkage applications where the inputs are an image of a text and the corresponding OCR, cross-modal attention is unlikely to lead to significantly richer representations.

We utilize the AdamW optimizer for all model training, combined with a Cosine Annealing with Warm Restarts learning rate (LR) scheduler. The maximum LR was set for each run, and the minimum LR was fixed at 0. The first restart was scheduled after 10 steps, with a restart factor of 2, doubling the time to restart after each cycle. Each epoch involved sampling $m$ views (image-text pairs) of each label in the dataset and processing them once.

The total training time was 34.6 hours on a single A6000 GPU card: 24 hours for language-image pretraining, 10 hours for supervised training using synthetic data, and 40 minutes for training on the hand-labeled data. Additional hyperparameters and model training details are provided in the supplemental materials.

A single A6000 GPU card accommodates a batch size \( B \) of 153 image-text pairs. In comparison, the original CLIP model \cite{radford2021learning} was trained with a batch size of 32,768 using 256 V100 GPUs. To ensure adequate in-batch negatives with the smaller batch sizes suitable for various downstream users, we employed offline hard negative mining. Details on batching and hard negative mining are detailed in the supplementary materials.  %All training was done on an A6000 40 GB GPU.  

At inference time, instances are linked using nearest neighbor retrieval. Facebook Artificial Intelligence Similarly Search (FAISS) \cite{johnson2019billion}, with \texttt{IndexFlatIP}, is used to calculate pairwise exact distances between the embeddings, in order to retrieve the nearest neighbor of each query (firm in the supplier lists) in the firm directories.\footnote{Because FAISS range search is not GPU-supported, we implement $k$ nearest neighbor search, conservatively setting $k$ to 900.}

Blocking - which incorporates type information from structured databases into record linkage - is important in many applications and has been the subject of extensive research (see \citet{steorts2014comparison, papadakis2019survey} for reviews). While not applicable to this study's applications, blocking is natural to incorporate into a contrastive setup, through using a type-specific loss function \cite{leszczynski2022tabi}. This will allow for matches to be made even when there is some noise in the type, a common scenario. 

To examine how \clippings compares to an analogous framework trained using a state-of-the-art unimodal encoder, we train a symmetric DINO vision transformer \cite{caron2021emerging} using the Japanese firm linked image crop data.
Details on self-supervised pre-training and hyperparameters for the self-supervised and supervised training are detailed in the supplementary materials. 

\section{Record Linkage} \label{linkage}

\subsection{Data}
This study's first application is constructing historical Japanese supply chains. 
This entails matching suppliers and customers recorded in firm level records collected in 1956  for over 7,000 large Japanese firms \cite{pr} to a firm level directory that provides additional rich information about nearly 70,000 firms \cite{teikoku}. The former are written horizontally and the latter vertically, making a potentially challenging case for visual matching. Firm name crops were localized using a Mask R-CNN \cite{he2017mask} model custom-trained with Layout Parser \cite{shen2021layoutparser}. 
To create labeled data for training and evaluation, the customers and suppliers of randomly selected firms were hand merged with the firm directory. Two annotators completed this task and resolved all discrepancies by hand. Many firms appear as customers and suppliers of multiple firms, and the data were de-duplicated such that each customer or supplier appears only once, in order to avoid leakage. Sometimes a single firm is linked to multiple entries in the firm directory, as firms can appear there more than once if they were registered in multiple prefectures. 

We used a 60-20-20 split to divide classes into a training set (772 examples), validation set (257 examples), and test set (257 examples). 
The test data links the customer-supplier list to all matching firms in the directory (with multiple matches occurring when a firm is registered in multiple prefectures), whereas this costly labeling was not needed for the training data, where each firm has a single match. In the main text, we report results using a dataset that drops customers and suppliers like ``the government'' that do not appear in the firm directory. In the supplementary materials, we report analyses including these instances, with similar patterns emerging. 

We also trained on 19,793 synthetically generated Japanese placenames, with an 80-20 train-val split. Each are rendered using different fonts and OCRed with two different OCR engines, that make different errors \cite{carlson2023, du2022svtr}. Each epoch involved sampling 3 "views" of each image crop-ocr pair.

Additionally, we conducted self-supervised language-image pre-training of the Japanese CLIP encoders, using 111,750 firm image crops and their corresponding OCR, as well as the same 19,793 synthetically generated names and their OCR, with an 80-20 test-val split.

As record linkage accuracy with string matching is likely to relate to the quality of the OCR, we apply string matching comparisons using two different OCRs of the Japanese firm names. The "noisy" dataset is created by using Google Cloud Vision off-the-shelf, as off-the-shelf OCR usage is the overwhelming norm (Google Cloud Vision does not support fine-tuning) and is often noisy. The "clean" dataset is created using a custom-trained OCR that achieves a character error rate of 0.6\% \cite{carlson2023}, a near best case scenario. Nevertheless, in character-based languages entities often have relatively few characters in their names, meaning even a single OCR error can destroy significant information.

\subsection{Results}

\begin{table}[t]
    \centering
    \resizebox{\linewidth}{!}{
    \begin{threeparttable}
       \begin{tabular}{lcc}
      \toprule
 & Noisy & Clean \\
	& OCR & OCR \\
\cmidrule{1-3}

\multicolumn{3}{l}{\textbf{\textit{Panel A: String-Matching}}} \\
Levenshtein distance & 0.630 & 0.731 \\
%Jaccard similarity &   \\
Stroke $n$-gram similarity  & 0.689 & 0.731 \\ \\
\multicolumn{3}{l}{\textbf{\textit{Panel B: Language-Image Self-Supervised Training}}} \\
Visual Linking & 0.769 & 0.769 \\
Language Linking & 0.740 & 0.790 \\
Multimodal Linking & 0.845 & 0.849 \\ \\
\multicolumn{3}{l}{\textbf{\textit{Panel C: Supervised Training on Linked Data}}} \\
\multicolumn{3}{l}{\textbf{\textit{\ \ with Vision Pre-training}}} \\
Visual Linking & 0.878 & 0.878 \\ \\
\multicolumn{3}{l}{\textbf{\textit{Panel D: Supervised Training on Linked Data}}} \\
\multicolumn{3}{l}{\textbf{\textit{\ \ with Language-Image Pre-training}}} \\
Visual Linking & 0.924 & 0.924 \\
Language Linking & 0.790 & 0.882 \\
Multimodal Linking  & 0.937 & \textbf{0.945} \\
      \bottomrule 
    \end{tabular}
    \end{threeparttable}
  }
    \caption{Baseline Matching Results: This table reports accuracy on the test set using a variety of different methods for linking Japanese firms from supply chain records to a large firm directory. \textit{Noisy OCR} uses off-the-shelf Google Cloud Vision for OCR, and \textit{Clean OCR} uses an accurate custom-trained OCR.}
      \label{matching_results}
\end{table}

\clippings achieves a record linkage accuracy of 94.5\% (Table \ref{matching_results}, Panel D), significantly outperforming string matching metrics (Panel A) on both noisy OCR (68.9\% accuracy) and clean OCR (73.1\% accuracy). When using multimodal representations, the accuracy of the OCR has only a very modest impact on linking accuracy: 93.7\% with noisy OCR versus 94.5\% with clean OCR. 

Using the pooled representations also outperforms tuning only the pre-trained image or pre-trained language encoders on paired entity data for that modality, which achieve an accuracy of 92.4\% (vision matching) and 88.2\% (language matching). 
When there are OCR errors, the image crops are useful as they avoid the destruction of information through OCR. At the same time, language understanding is helpful due to abbreviations and multiple ways to write a firm name. 

To this end, using a vision encoder that was multimodally pre-trained on image crop-OCR'ed text paired data achieves an accuracy of 92.4\%, beating the unimodally trained vision encoder - which achieves an accuracy of 87.8\% - by a fairly wide margin (Panel C). This suggests the utility of multimodal pre-training, which we will examine further in the error analysis. 

\begin{figure*}
    \centering
    \includegraphics[width=.8\linewidth]{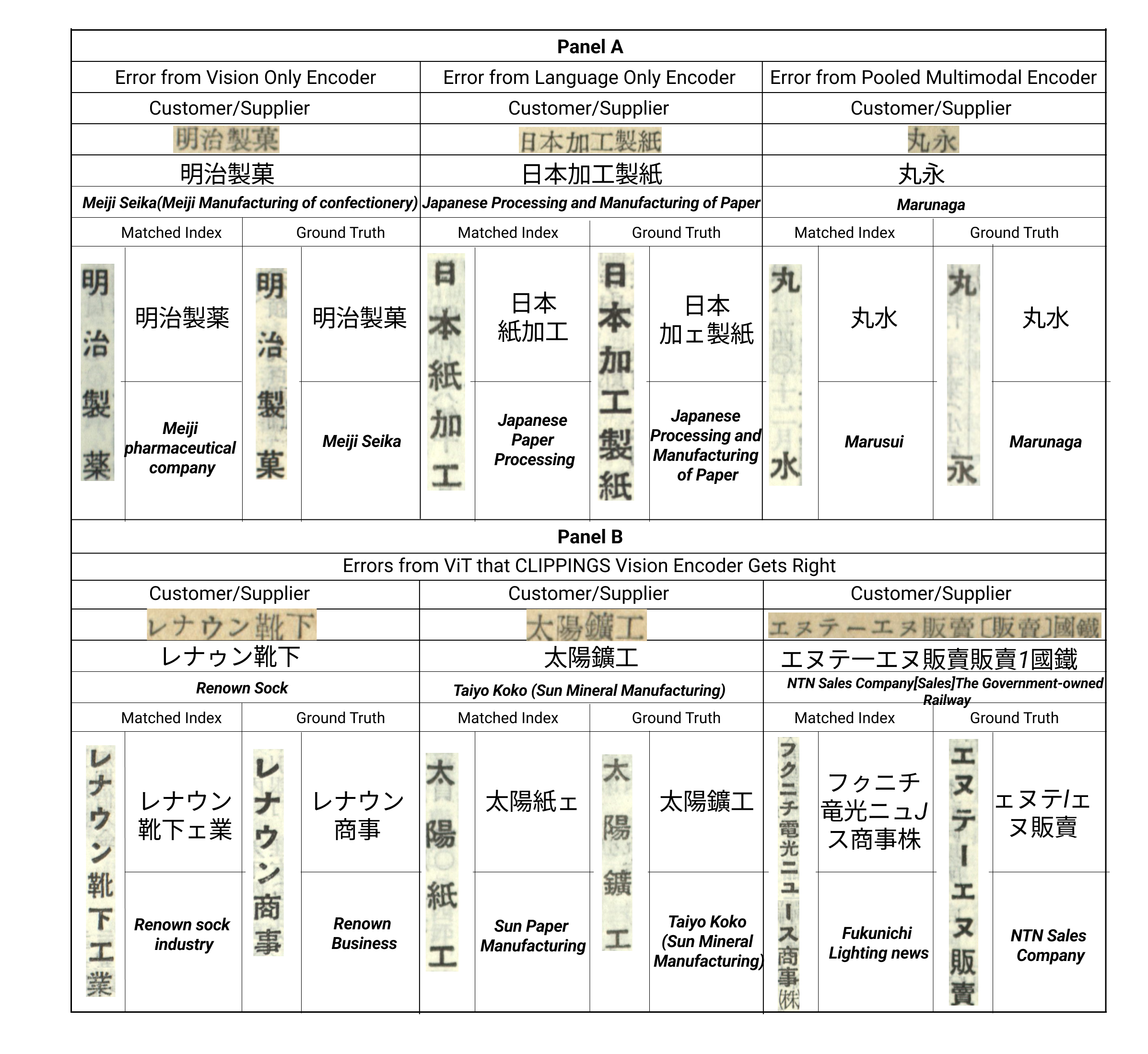}
    \caption{Errors: This figure shows errors in record linkage made by different models.} 
    \label{fig:error}
    \vspace{-4mm}
  \end{figure*}

Figure \ref{fig:error}, Panel A shows representative errors made by the supervised vision only, language only, and multimodal encoders, that use language-image pre-training (Panel D).  
In the vision-encoder error, the ground truth is Meiji Seika, a firm producing snacks and chocolates.  Its prediction is Meiji pharmaceutical company, as 菓 and 薬 look similar. This illustrates how the vision encoder can sometimes confuse visually similar entities that have very different meanings.
In the language-only encoder error, the ground truth is ``Japanese Processing and Manufacturing of  Paper''（日本加工製紙), and the prediction is ``Japanese Paper Processing''（日本紙加工), very similar in meaning. The multimodal encoder predicts both of these cases correctly.
Finally, the multimodal error is a company whose name is only two characters. The ground truth is 丸永, and the prediction is 丸水, with 永 and 水 very visually similar. These are names without language context, so the language encoder cannot distinguish them either.

Fascinatingly, the tuned vision only encoder with language-image pre-training gets some matches that require language understanding correct (accuracy 92.4\%), that the tuned ViT model with vision-only pre-training gets wrong (accuracy 87.8\%), suggesting some acquisition of language understanding from the multimodal pre-training.\footnote{Tuning a Japanese S-BERT language bi-encoder does even worse than the vision only encoder, but is less of an apples-to-apples comparisons to the \clippings text-only model due to the challenges of devising self-supervised language-only pre-training recipes in this context.}
Figure \ref{fig:error}, Panel B provides several examples. 
In the first example, レナウン靴下（Renown Sock) is matched by the ViT model to “レナウン商事(Renown Business), whereas the multimodally pre-trained encoder matches it correctly to レナウン靴下工業（Renown Sock Industry), despite the extra two characters.
In the second example, the ground truth is 太陽鑛工 (Sun Mineral Manufacturing), whereas the ViT prediction is 太陽紙工（Sun Paper Manufacturing).
The third example results from an error in our custom trained layout detection model, that concatenates two customer-suppliers. The detected firm is ``エヌティエヌ販賣會社[販賣]国鉄,'' which translates as ``NTN Sales Company[Sales]The Government-owned Railway''. The ViT  simply predicts a company of similar length, whereas the encoder with multimodal pre-training matches it to NTN Sales Company. 
%These examples show that the vision encoder gains language understanding through the contrastive language-image self-supervised pre-training. 

The purely self-supervised multimodal model also outperforms the string matching methods, with an accuracy of 84.9\% (Table \ref{matching_results}, Panel B). $n$-gram similarity at the stroke level with fuzzychinese \cite{fuzzychinese} achieves an accuracy of 73.1\% on clean OCR, as coincidentally does Levenshtein distance. When only the vision encoder or language encoder of the self-supervised multimodal model is used at test time, the performance also exceeds that of standard string matching techniques (76.9\% and 79.0\% accuracy, respectively). This shows that with only self-supervised training, neural methods can still outperform widely used string matching methods. 

\begin{table}[ht]
    \centering
       % \resizebox{\textwidth}{!}{

       \begin{tabular}{lcc}
      \toprule
 & Noisy OCR & Clean OCR \\
\cmidrule{1-3}

\multicolumn{3}{l}{\textbf{\textit{Panel A: Zero-shot Japanese-CLIP}}} \\
Visual Linking & 0.000 & 0.000 \\
Language Linking & 0.639 & 0.626 \\
Multimodal Linking  & 0.491 & 0.433 \\
\multicolumn{3}{l}{\textbf{\textit{Panel B: Only Supervised Training }}} \\
Multimodal Linking  & 0.676 & 0.731 \\

      \bottomrule 
    \end{tabular} %}
    \caption{Record Linkage Ablations: This table reports accuracy for linking Japanese firms from supply chain records to a large firm directory. \textit{Noisy OCR} uses off-the-shelf Google Cloud Vision for OCR, and \textit{Clean OCR} uses an accurate custom-trained OCR. Panel A uses Japanese CLIP off-the-shelf, and Panel B uses only supervised training, without further self-supervised pre-training.}
      \label{add_matching_results}
\end{table}

Table \ref{add_matching_results} examines the contribution of different elements of the \clippings training recipe. 
Panel A considers the performance of Japanese CLIP \cite{japanese-clip} off-the-shelf. Using the vision encoder alone, every instance is mispredicted. The off-the-shelf text encoder, while better than the vision encoder, is outperformed by traditional string matching methods. 

\begin{figure*}[ht]
    \centering
    \includegraphics[width=.8\linewidth]{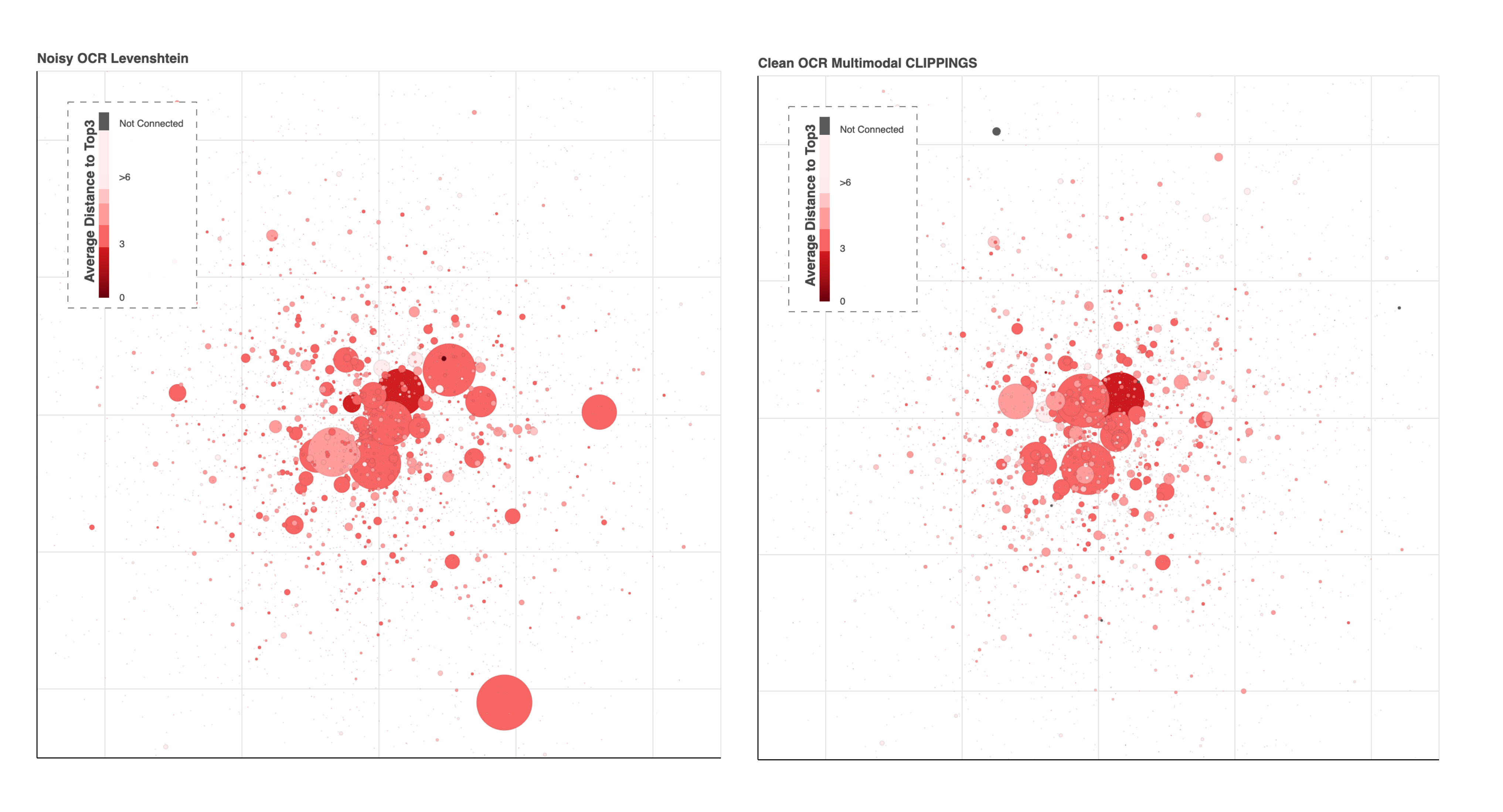}
    \caption{Input-Output Networks: This figure plots the average supply chain distance of Japanese firms to Mitsui, Mitsubishi, and Sumitomo, the three most prominent Japanese firms.} 
    \label{fig:output}
    \vspace{-4mm}
  \end{figure*}

Panel B examines the performance of \clippings when only supervised training is used, discarding the self-supervised training on image-OCR pairs. Performance declines significantly relative to when self-supervised language-image pre-training is employed - with accuracy similar to that of traditional string matching methods - illustrating the importance of first aligning the vision and text encoders for the document image-OCR domain.

Since \clippings leverages transfer learning from language-image pre-training, it can be precisely tailored to different settings with small, cheap-to-create training sets, or used as a purely self-supervised solution. 
Moreover, hard negatives in contrastive training encourage separation between different instance representations even if the instances are very similar, making it highly suitable for linking datasets with lots of confusables. 
%Our application does not incorporate blocking, as it is not necessary in this setting, but in the many settings where it is we conjecture that the expressiveness of a type-specific contrastive loss could be another benefit. 

Deployment costs are important for extensibility, as record linkage problems often entail linking millions of entities on a limited compute budget. Experiments on the full dataset of 36,673 customer-suppliers, using an 18 core Intel(R) i9-9980XE CPU @ 3.00GHz and a single NVIDIA A6000 GPU, show that \texttt{CLIPPING}'s deployment costs are modest, making it a highly scalable solution. 
Mutlimodal \clippings takes 6 minutes, 21 seconds to run on the full data with a single GPU, all but one second of which are embedding the crops and OCR. This compares to 
54 minutes to implement our optimized CPU Levenshtein distance calculations (which is an order of magnitude faster than R string matching, \textit{e.g.} \cite{Rpackage}) and 3 minutes and 7 seconds for stroke matching with fuzzychinese \cite{fuzzychinese}.

Figure \ref{fig:output} illustrates the supply chain networks created by applying \clippings to the full dataset, with the shading showing the average distance in the supply chain networks to the three largest Japanese conglomerates: Mitsui, Mitsubishi, and Sumitomo. Node position and scaling are fixed across the graphs, with scaling showing average degree centrality in the supply chain network. The key takeaway is that using off-the-shelf OCR and Levenshtein distance - a prevalent approach in the literature - creates a visibly different network (left) than the multimodal method (right), which is much closer to the ground truth. 
For example, in the bottom right corner of the graph created with edit distance, there is a firm with lots of false links (and hence a high degree centrality).
%7,352 firm nodes (56\%) are in the supply chains of these big-3 firms.  
A study of the Japanese economy based on the noisier network is likely to produce biased results \cite{chandrasekhar2011econometrics}. 
This illustrates the utility for downstream social science research of improving record linkage through multimodal learning. 

\clearpage

\section{Limitations} \label{limits}

The biggest limitation of \clippings is that it is currently tuned for a particular language and type of document collection. Because strong performance can be achieved even with purely self-supervised training, supervised tuning does not require many labels, and the compute requirements are not too onerous, it would be relatively straightforward to extend to a much more diverse, multilingual document collection. We hope that by illustrating the utility of this method, that it will encourage further development of multimodal models for record linkage.

\section{Ethical Considerations} \label{ethics}
\clippings is ethically sound. Its
methods are entirely open source, and its training
data are entirely in the public domain. While \clippings is more accurate than string matching, as well as unimodal neural matching, some errors will often still arise with multimodal matching. It is important for researchers to assess whether these could affect their conclusions. 

%% Insert supplemental materials for arxiv

\clearpage
\begin{center}
    \section*{Supplementary Materials}

\end{center}

% \maketitle

\section{Methods}

\subsection{Japanese Multimodal Models}
The Japanese multimodal models were initialized with a Japanese CLIP checkpoint \cite{japanese-clip}. Japanese CLIP was trained with the standard CLIP \cite{radford2021learning} loss but used a BERT-based text encoder and the vision transformer was initialized by weights from the AugReg ViT-B/16 model \cite{augreg}.

\subsubsection*{Synthetic Data}
Both language-image pretraining and the supervised training of \clippings employed synthetic data. To create synthetic data, we rendered a list of common Japanese words as images using different fonts (a type of augmentation), applied image augmentations, and fed the resulting images to OCR. For the same word, this produced varying views of the word's image due to different augmentations, as well as different views of the word's text due to varying OCR errors induced by the augmentations. 

We randomly sample one image-text pair per label (word) and use this subset for language-image pretraining. For training \clippings, we train on the full synthetic dataset and then fine-tune on our labelled data.

\subsubsection*{Other Training Details}

Text crops are thin vertically or horizontally oriented rectangles, with highly diverse aspect ratios, whereas vision encoders are almost always trained on square images. Center cropping would discard important information and resizing often badly morphs the image, given that the underlying aspect ratios are far from a square. To preserve the aspect ratio, we pad the rectangular crop with the median value of the border pixel such that the text region is always centered. 

For language-image pretraining, the standard CLIP loss was used to align the image and text encoders \cite{radford2021learning}. Supervised Contrastive loss \cite{khosla2020supervised} was used for all supervised training. We used the AdamW optimizer for all model training along with a Cosine Annealing with Warm Restarts learning rate (LR) scheduler where the maximum LR was specified for each run and the minimum LR was set to 0. 10 steps were chosen for the first restart with a restart factor of 2 that doubles the time to restart after every restart. An epoch involved sampling $m$ views (image-text pair) of each label in the dataset and going through each of them once. It took a 24 hours to perform language-image pretraining, 10 hours to perform supervised training using synthetic data and 40 minutes to train on the labelled data - a total training time of 34.6 hours to train \clippings on a single A6000 GPU card. 
Hyperparameters and additional details about model training are listed in Table \ref{tab:rl_hp}.

At inference time, we used  \textit{IndexIPFlat} from FAISS \cite{johnson2019billion} to find the nearest neighbor on L2-normalized embeddings.

\subsubsection*{Hard Negative Mining}
\clippings was trained on a single A6000 GPU card, which could fit a batch size $B$ of 153 image-text pairs. This compares to a batch size of 32,768 used to train the original CLIP \cite{radford2021learning} on 256 V100 GPUs. Offline hard negative mining was used to achieve sufficient in-batch negatives with the small batch sizes that can fit into compute setups that are realistic for diverse downstream users. 

Define the data as a triple $(x_n,t_n,y_n)$,
where $x_n \in \mathcal{X}$ is the image, $t_n \in \mathcal{T}$ is the text, and $y_n \in \mathcal{Y}$ is an associated label (class). Let $D$ be the set of all triples. 
For supervised pertaining using synthetic data, we randomly sampled one image-text pair per label in $D$ to form $D' \subset D$. For each image-text pair $(x_a,t_a)$ in $D'$, we use the domain-adapted CLIP model to find its $k$  nearest neighbor pairs $(x_k,t_k)$ (including itself). This gives us the $k-1$ nearest neighbours for each label $y_a$ in $D' \subset D$ .

The anchor label $y_a$ and its $k-1$ neighbors form a "hard-negative" set. In a batch, we sample $m=3$ views of image-text pairs with replacement. A batch-size divisible by $k*m$ can fit $\dfrac{B}{k*m}$ unique classes, each with their own $m$ views and the $m$ views of all $k$ neighbors. 
We shuffle the hard-negative sets and partition them into groups of $\dfrac{B}{k*m}$ such that each group can constitute a batch. Each constituent class has $k$ neighbors and $m$ views within the minibatch.
For the next step - fine-tuning with labeled data - we follow the same approach but with the best model checkpoint from synthetic pretraining.

% \subsection{English Multimodal Models}

% Our English \clippings uses the official OpenAI CLIP model \cite{radford2021learning}, which has a ViT-B/32 Transformer architecture for the image encoder and a masked self-attention Transformer as the text encoder. 
% To train \clippings for this application, we used the standard image processor, which resized on the shortest edge, center cropped, and normalized. Language-image pretraining took 13 hours and the supervised training on labeled image-text pairs took 18 hours on a single NVIDIA GeForce RTX 3090. Thus, it took a total of 21 hours to train english multimodal clippings.
% Hyperparameters and other training details are listed in Table \ref{tab:rl_hp}.

% To cluster the embeddings, we use Single Linkage Clustering. The distance threshold was tuned on the validation set jointly with the weight put on the image/text embedding to create the pooled embedding. 

\subsection{Vision Transformer}

We initialized the weights of the Vision Transformer from the DINO-pretrained checkpoint for ViT/B16 \cite{caron2021emerging}. 
Hyperparameters and other training details are listed in Table \ref{tab:rl_hp}.

As for \clippings, ViT training employed synthetic data. The same pipeline as above was used to generate synthetically noised images. For each word in a list of Japanese words, the text was rendered using different fonts and augmented to create synthetically noised views. 

Offline hard-negative mining was used to train the ViT. The approach is similar to that described above. We used a pretrained checkpoint to find $k$ nearest neighbors for each class.  This was used to create a batch containing $m$ views of that class, along with $m$ views of $\dfrac{B}{m} - 1$ other classes. When using hard negatives, we substituted "k-1" of these other classes with the nearest neighbor classes of the anchor.

\section{Additional Results}

Table \ref{nomatch_results} examines the inclusion of instances without a match in the firm directory, for example, a range of government agencies. While performance declines somewhat relative to the results reported in the main text, the relative comparisons between models hold.

\clearpage

\setcounter{table}{0}
\renewcommand{\thetable}{S-\arabic{table}} % Setting the table number output to letters 
\setcounter{figure}{0}
\renewcommand{\thefigure}{S-\arabic{figure}} % Setting the figure number output to letters 

\onecolumn

\begin{table}[ht]
\centering

\begin{tabular}{lccccccccc}
\toprule \\

\multicolumn{1}{c}{\textbf{Model}} &
  \multicolumn{1}{c}{\textbf{lr}} &
  \multicolumn{1}{c}{\textbf{B}} &
  \multicolumn{1}{c}{\textbf{w\_decay}} &
  \multicolumn{1}{c}{\textbf{temp}} &
  \multicolumn{1}{c}{\textbf{im\_wt}} &
  \multicolumn{1}{c}{\textbf{m}} &
  \multicolumn{1}{c}{\textbf{k}} &
  \multicolumn{1}{c}{\textbf{epochs}} &
   \multicolumn{1}{c}{\textbf{nm\_thresh}} 
   \\
  \midrule
Language-image Pretraining          & 5e-5   & 153 & 0.001  & 0.048 & -   & - & - & 40 & - \\
\textit{Supervised models} &        &     &        &       &     &   &  & &   \\
ViT (synthetic)                     & 5.8e-5 & 256 & 0.0398 & 0.048 & -   & 8 & 8  & 5 &  - \\
ViT (labelled)                      & 2e-6   & 252 & 0.1    & 0.09  & -   & 3 & 8  & 10 & 0.88 \\
Sup. Lang.-only (synthetic)         & 5e-6   & 153 & 0.001  & 0.1   & 0   & 3 & 3  & 30 & 0.85, 0.85 \\
Sup. Image-only (synthetic)         & 5e-6   & 153 & 0.001  & 0.1   & 1   & 3 & 3  & 30 & 0.76 \\
Sup. Mean-pool (synthetic)          & 5e-6   & 153 & 0.001  & 0.1   & 0.5 & 3 & 3  & 30 & 0.81, 0.80 \\
Sup. Lang-only (labelled)          & 5e-6   & 153 & 0.001  & 0.1   & 0   & 3 & 3  & 30 & 0.84,0.82 \\
Sup. Image-only (labelled)          & 5e-6   & 153 & 0.001  & 0.1   & 1   & 3 & 3  & 30 & 0.79 \\
Sup. Mean-pooling (labelled)        & 5e-6   & 153 & 0.001  & 0.1   & 0.5 & 3 & 3  & 30 & 0.82,0.82 \\ 
\bottomrule 
\end{tabular}
\caption{Training Hyperparameters: $lr$ is the maximum learning rate, $B$ is the batch size, $w\_decay$ is the AdamW weight decay,  $im\_wt$ is the weight of the image embedding in the pooled embedding, $m$ is the number of views sampled in each epoch,  $k$ is the number of nearest neighbours in a hard-negative set, and epochs is the number of epochs. $nm\_threshold$ is a tuple with two tuned similarity thresholds (for noisy and clean OCR respectively) under which a retrieved neighbor is considered to not match with any of the target images.}
\label{tab:rl_hp}
\end{table}

\clearpage
\begin{table}[ht]
    \centering
       % \resizebox{\textwidth}{!}{

       \begin{tabular}{lcc}
      \toprule
 & Noisy OCR & Clean OCR \\
\cmidrule{1-3}

\multicolumn{3}{l}{\textbf{\textit{Panel A: String-Matching}}} \\
Levenshtein distance & 0.605 & 0.625 \\
%Jaccard similarity &   \\
Stroke $n$-gram similarity  & 0.625 & 0.650 \\ \\
\multicolumn{3}{l}{\textbf{\textit{Panel B: Language-Image Self-Supervised Training}}} \\
Visual Linking & 0.693 & 0.693 \\
Language Linking & 0.680 & 0.741 \\
Multimodal Linking & 0.770 & 0.770 \\ \\
\multicolumn{3}{l}{\textbf{\textit{Panel C: Supervised Training on Linked Data}}} \\
\multicolumn{3}{l}{\textbf{\textit{\ \ with Vision Pre-training}}} \\
Visual Linking & 0.819 & 0.819 \\ \\
\multicolumn{3}{l}{\textbf{\textit{Panel D: Supervised Training on Linked Data}}} \\
\multicolumn{3}{l}{\textbf{\textit{\ \ with Language-Image Pre-training}}} \\
Visual Linking & 0.829 & 0.829 \\
Language Linking & 0.757 & 0.825 \\
Multimodal Linking  & 0.845 & \textbf{0.871} \\
      \bottomrule 
    \end{tabular}
    \caption{Including instances without a match: This table reports accuracy on the test set using a variety of different methods for linking Japanese firms from supply chain records to a large firm directory. \textit{Noisy OCR} uses off-the-shelf Google Cloud Vision for OCR, and \textit{Clean OCR} uses an accurate custom-trained OCR.}
      \label{nomatch_results}
\end{table}

\clearpage

\bibliography{cites}

\end{document}